\documentclass[twoside,11pt]{article}

\usepackage[abbrvbib]{jmlr2e}

\usepackage[utf8]{inputenc} 
\usepackage[T1]{fontenc}    
\usepackage{microtype}      

\usepackage{float}
\usepackage{enumitem}
\usepackage{tcolorbox}
\usepackage{graphicx}      
\usepackage{subcaption}   

\usepackage[capitalise,noabbrev]{cleveref}

\usepackage{listings}
\usepackage{adjustbox}
\usepackage{pythonhighlight}
\usepackage{xcolor} 
\usepackage[frozencache,cachedir=.]{minted}
\setminted{escapeinside=||}
\tcbuselibrary{minted}

\usepackage{lastpage}

\ShortHeadings{ObjectRL: An Object-Oriented Reinforcement Learning Codebase}{Baykal, Akg\"ul, Haussmann, Tasdighi, Werge, Wu and Kandemir}
\firstpageno{1}

\begin{document}

\title{ObjectRL: An Object-Oriented Reinforcement Learning Codebase}

\author{\name Gulcin Baykal\thanks{Corresponding author} \email baykalg@imada.sdu.dk 
       \vspace{-0.15cm}
       \AND
       \name Abdullah Akg\"ul \email akgul@imada.sdu.dk
       \vspace{-0.15cm}
       \AND
       \name Manuel Haussmann \email haussmann@imada.sdu.dk 
       \vspace{-0.15cm}
       \AND
       \name Bahareh Tasdighi \email tasdighi@imada.sdu.dk
       \vspace{-0.15cm}
       \AND
       \name Nicklas Werge \email werge@sdu.dk
       \vspace{-0.15cm}
       \AND
       \name Yi-Shan Wu \email yswu@imada.sdu.dk
       \vspace{-0.15cm}
       \AND
       \name Melih Kandemir \email kandemir@imada.sdu.dk
       \AND
       \addr Department of Mathematics and Computer Science \\
       University of Southern Denmark \\
       Odense, Denmark
}

\maketitle

\begin{abstract}
\texttt{ObjectRL} is an open-source Python codebase for deep reinforcement learning (RL), designed for research-oriented prototyping with minimal programming effort. Unlike existing codebases, \texttt{ObjectRL} is built on Object-Oriented Programming (OOP) principles, providing a clear structure that simplifies the implementation, modification, and evaluation of new algorithms. \texttt{ObjectRL} lowers the entry barrier for deep RL research by organizing best practices into explicit, clearly separated components, making them easier to understand and adapt. Each algorithmic component is a class with attributes that describe key RL concepts and methods that intuitively reflect their interactions. The class hierarchy closely follows common ontological relationships, enabling data encapsulation, inheritance, and polymorphism, which are core features of OOP. We demonstrate the efficiency of \texttt{ObjectRL}'s design through representative use cases that highlight its flexibility and suitability for rapid prototyping. The documentation and source code are available at \href{https://objectrl.readthedocs.io}{\color{blue}\texttt{https://objectrl.readthedocs.io}} and \href{https://github.com/adinlab/objectrl}{\color{blue}\texttt{https://github.com/adinlab/objectrl}}.
\end{abstract}


\section{Introduction} \label{sec::introduction}

Deep reinforcement learning (RL) has led to advances in autonomous decision-making \citep{sutton2018reinforcement}, with applications ranging from video game playing and robotics to autonomous driving and large language models that incorporate human feedback \citep{mnih2015human, lillicrap2016continuous, schulman2017proximal, smith2023demonstrating, wang2023efficient, ramesh2024group}. A variety of open-source RL codebases have been developed to support algorithm development and benchmarking \citep{liang2018rllib, munoz2023skrl, huang2020cleanrl, raffin2021stable, weng2022tianshou, bou2023torchrl, zhu2024pearl, eschmann2024rltools, deramo2021mushroomrl}. We provide a comparison in~\cref{sec::related_work}. While many codebases support scalability, simplicity, and reproducibility, their tightly coupled components, complex configurations, and deep functional abstractions limit direct access to individual algorithmic elements and make them difficult to extend. This makes implementing and evaluating new ideas challenging, particularly for researchers new in the field.

To support algorithmic innovation, we present \texttt{ObjectRL}, the first Python codebase designed with a strong focus on object-oriented programming principles. Rather than providing yet another monolithic framework, \texttt{ObjectRL} offers a clear class structure that mirrors the conceptual building blocks of modern RL algorithms. Core components, such as agents, actor policies, critic ensembles and their target networks, are implemented as independent and reusable classes. This approach simplifies the implementation, modification, and comparison of RL algorithms. This isolation of design choices enables researchers to iterate and experiment with new ideas more efficiently.

\section{Related Work} \label{sec::related_work}

The growing popularity and practical impact of deep RL have led to the development of a wide range of open-source codebases. \texttt{RLlib} \citep{liang2018rllib}, \texttt{skrl} \citep{munoz2023skrl}, and \texttt{Pearl} \citep{zhu2024pearl} target production robustness and large-scale or robotics deployment. However, their complex APIs and configuration-heavy architectures can hinder rapid algorithmic prototyping and make it harder to trace or modify low-level components. \texttt{Stable-Baselines3} (\texttt{SB3}) \citep{raffin2021stable} and \texttt{CleanRL} \citep{huang2020cleanrl} focus on simple off-the-shelf usage and reproducibility. \texttt{CleanRL} follows a monolithic and single-file architecture, which simplifies readability but reduces code modularity. \texttt{SB3} provides stable and modular implementations, but its unstructured API makes navigation and customization harder. \texttt{Tianshou} \citep{weng2022tianshou} and \texttt{TorchRL} \citep{bou2023torchrl} offer strong modular designs but their abstraction layers and component integration can limit the intuitive prototyping of new algorithms. \texttt{RLTools} \citep{eschmann2024rltools} focuses on performance and composability with a functional design, which can hinder intuitive low-level modifications. \texttt{MushroomRL} offers a modular design for research-oriented implementation similarly to our purpose. However, it makes only partial use of the object-oriented design principles, which makes algorithmic exploration less straightforward than our proposal.

\texttt{ObjectRL} fully applies object-oriented design to individual algorithmic components, offering an intuitive class structure that prioritizes readability, debuggability, and ease of modification.  This design enables easier prototyping of new algorithms, especially when building on existing implementations.

\section{ObjectRL: Object-Oriented Design for Rapid Prototyping} \label{sec::objectrl}

\texttt{ObjectRL} provides clean implementations of some widely used RL algorithms, such as DQN \citep{mnih2015human}, DDPG \citep{lillicrap2016continuous}, PPO \citep{schulman2017proximal}, TD3 \citep{fujimoto2018addressing}, and SAC \citep{haarnoja2018soft}. These implementations serve as reliable baselines and as flexible starting points for developing and testing new ideas. The class structure enables rapid exploration of research ideas with minimal programming effort, making \texttt{ObjectRL} a valuable tool for research as well as education.

\paragraph{Why Object-Oriented Design for RL?}

Deep RL algorithms consist of multiple interacting components. Structuring these components in an extensible and well-separated way is critical for algorithmic research and prototyping. Object-Oriented Programming (OOP) is particularly well-suited for this task due to three key reasons:
\begin{itemize}[leftmargin=*, noitemsep, topsep=0pt]
    \item \textbf{Encapsulation}: OOP enforces a clear separation of responsibilities by encapsulating class attributes and allowing access only through method calls. In \texttt{ObjectRL} each RL component (e.g., agent, critic, policy) is represented by a dedicated class. This design enables easy debugging and modification without affecting the other parts of the agent.
    \item \textbf{Inheritance and Composition}: OOP connects entities through inheritance (e.g., specializing base classes such as a general \texttt{Actor} into \texttt{SACActor} or \texttt{PPOActor}) and composition (e.g., combining objects such as policy networks, value networks, data buffers, and loggers to form a complete agent), reflecting how RL algorithms are built from interacting parts.
    \item \textbf{Polymorphism}: OOP allows the entities to share a common interface while providing different implementations (e.g., different \texttt{update()} methods for various actor types). This enables new types of behavior (e.g., alternative target updates, custom exploration strategies) to be added by extending or overriding components without modifying the backbone class structure.
\end{itemize}

\paragraph{Class Structure: A Case Study of SAC.}

At the core of \texttt{ObjectRL} lies an intuitive object-oriented decomposition of RL components into a structured class ontology. The central element is the abstract \texttt{Agent} class, which defines common methods for learning and environment interaction while holding common utilities such as \texttt{replay buffers} and \texttt{loggers}. Building on this, \texttt{ActorCritic} class defines the high-level structure of actor-critic methods. The \texttt{Model} class inherits \texttt{ActorCritic} and implements algorithm-specific details. 

\begin{figure}[t]
    \centering
    \includegraphics[width=\textwidth]{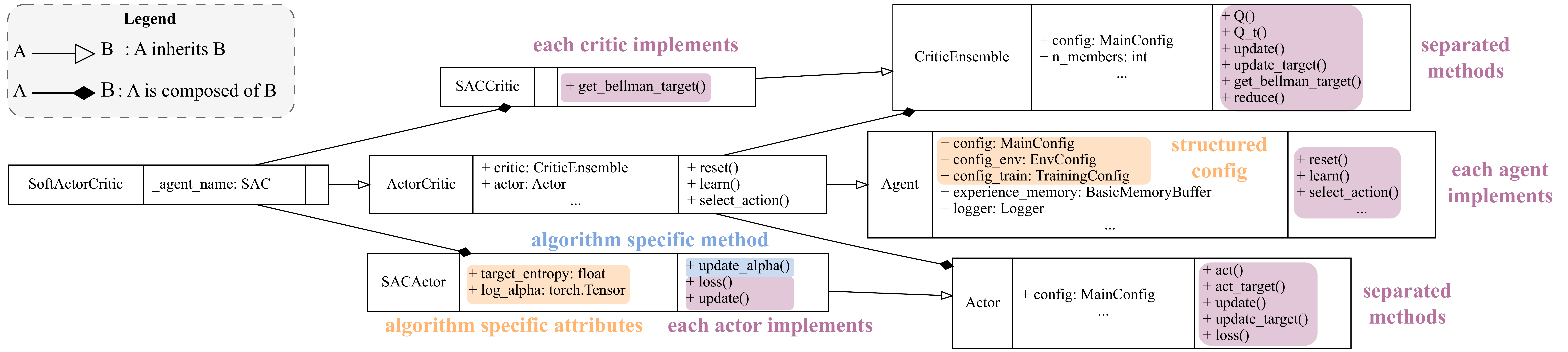}
    \caption{The class diagram of a Soft Actor-Critic implementation in the \texttt{ObjectRL} codebase. Core components and subclasses are highlighted. Inheritance and composition are shown with arrows and diamonds; key attributes, configurations, and methods are color-coded.}
    \label{fig:design}
\end{figure}

\Cref{fig:design} illustrates the class structure of an implementation of the SAC in \texttt{ObjectRL}. The \texttt{SoftActorCritic} agent is composed of a \texttt{SACActor} and a \texttt{SACCritic} instance, both inheriting from base classes defining shared behavior. The critic is modeled as a \texttt{CriticEnsemble}, which combines multiple critic networks and uses ensemble aggregation strategies like min-clipping. Algorithmic operations, such as deteministic or stochastic action generation, value or Bellman target estimation, and loss computation are implemented as methods in each class. This design allows for easy extension or replacement of individual elements (e.g., adding target smoothing or ensemble aggregation) with small localized changes.

\texttt{ObjectRL}'s module hierarchy is designed in an interpretable manner: \texttt{agents} provide base agent classes; \texttt{models} define actors, critics, their compositions, and algorithms; \texttt{config} manages hyperparameters via Python data classes; \texttt{experiments} handle training and evaluation; \texttt{loggers} support systematic result tracking; \texttt{nets} specify policy and value network definitions; \texttt{replay buffers} store experiences; and \texttt{utils} offer helper functions and tools.

\paragraph{From SAC to DRND: A Prototyping Example.}

We illustrate how \texttt{ObjectRL} can facilitate the extension of SAC into the framework of DRND \citep{yang2024exploration}. DRND augments SAC by injecting a bonus term into the actor and critic losses. This term encourages exploration in uncertain state-action regions, guided by disagreement across $Q$-value estimates. Implementing this algorithmic extension in \texttt{ObjectRL} is straightforward:
\begin{enumerate}[leftmargin=*, noitemsep, topsep=3pt]
    \item Define a new class, \texttt{DRNDBonus}, that encapsulates the ensemble-based uncertainty estimation and its associated hyperparameters.
    \item Extend the SAC actor and critic by creating \texttt{DRNDActor} and \texttt{DRNDCritic} classes that incorporate the exploration bonus into their \texttt{loss()} computations using polymorphism.
\end{enumerate}

The actor loss is extended trivially by overriding its \texttt{loss()} method to include the exploration bonus alongside the standard entropy-regularized objective inherited from \texttt{SACActor}:

\begin{center}
\begin{tcolorbox}[
    colback=white!99!blue,
    colframe=blue!20!black,
    arc=3pt,
    boxrule=0.5pt,
    left=3pt,
    right=3pt,
    top=3pt,
    bottom=3pt,
    width=0.6\linewidth,
]
\begin{minted}[fontsize=\footnotesize]{python}
loss, act_dict = super().loss(state, critics)
action = act_dict["action"]
|\colorbox{orange!30}{bonus = bonus\_ensemble.bonus(state, action).mean()}|
return loss |\colorbox{orange!30}{+ bonus}|, act_dict
\end{minted}
\end{tcolorbox}
\end{center}
To incorporate the bonus into the critic's Bellman target calculation, \texttt{DRNDCritic} overrides the base class method \texttt{get\_bellman\_target}, which computes the target $Q$-value. In this context, \texttt{target\_reduced} refers to the standard Bellman backup value computed by aggregating target critic predictions, excluding the exploration bonus and entropy terms. The modified method adds the exploration bonus to this target as follows:

\begin{center}
\begin{tcolorbox}[
    colback=white!99!blue,
    colframe=blue!20!black,
    arc=3pt,
    boxrule=0.5pt,
    left=3pt,
    right=3pt,
    top=3pt,
    bottom=3pt,
    width=0.9\linewidth,
]
\begin{minted}[fontsize=\footnotesize]{python}
|\colorbox{orange!30}{bonus = bonus\_ensemble.bonus(next\_state, next\_action)}|
q_target = target_reduced - alpha * log_prob |\colorbox{orange!30}{- self.lambda\_critic * bonus}|
return q_target
\end{minted}
\end{tcolorbox}
\end{center}

Our class ontology enables a simple implementation of SAC, making its extension to DRND straightforward as highlighted above. \texttt{ObjectRL}'s inheritance preserves compatibility with the existing training loop, requiring only minor additions to update the bonus predictors. This example illustrates the \texttt{ObjectRL}'s core goal: enabling researchers to prototype, evaluate, and debug new ideas. Additional examples are available in the documentation.\footnote{\url{https://objectrl.readthedocs.io/en/latest/examples.html}}



\acks{GB, AA, MH, and BT thank the Carlsberg Foundation (grant number CF21-0250); NW and YSW thank the Novo Nordisk Foundation (grant number NNF-21OC0070621) for supporting their research. We thank Ugurcan Ozalp for his contributions to the early stages of the development process.} 


\newpage

\appendix

\section{Results on Standard Benchmarks}

In addition to the implementations of DQN \citep{mnih2015human}, DDPG \citep{lillicrap2016continuous}, PPO \citep{schulman2017proximal}, TD3 \citep{fujimoto2018addressing}, and SAC \citep{haarnoja2018soft}, we also include recent directed exploration methods such as OAC \citep{ciosek2019better}, REDQ \citep{chen2021randomized}, DRND \citep{yang2024exploration}, and PBAC \citep{tasdighi2024deep}. These algorithms integrate seamlessly within \texttt{ObjectRL}'s class structure, enabling reuse and extension of core RL components like actors, critics, and update rules with minimal effort. This highlights \texttt{ObjectRL}'s flexibility for rapid prototyping and experimentation.

\begin{figure}[!th]
    \centering
    \begin{subfigure}[b]{0.45\textwidth}
        \centering
        \includegraphics[width=\textwidth]{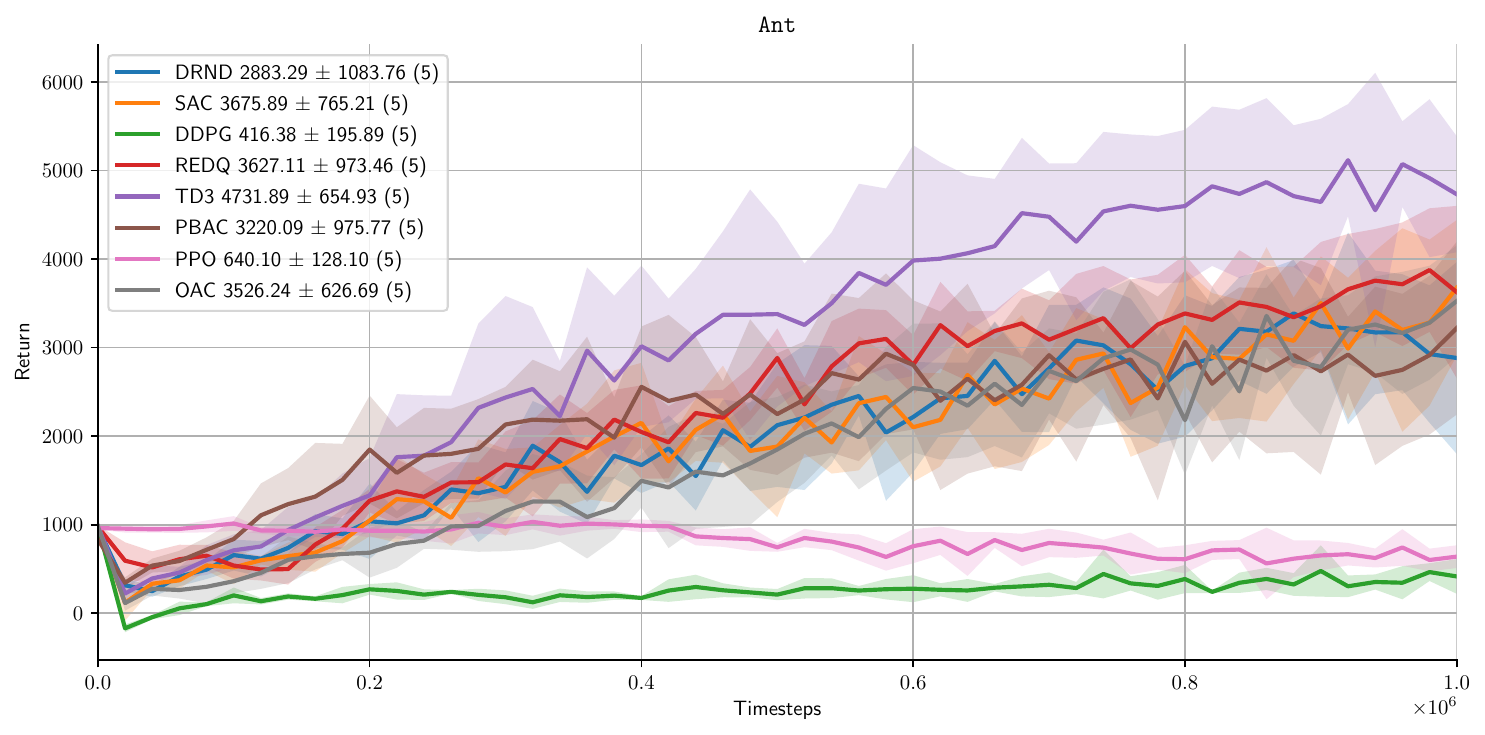}
        \label{fig:ant}
    \end{subfigure}
    \hfill
    \begin{subfigure}[b]{0.45\textwidth}
        \centering
        \includegraphics[width=\textwidth]{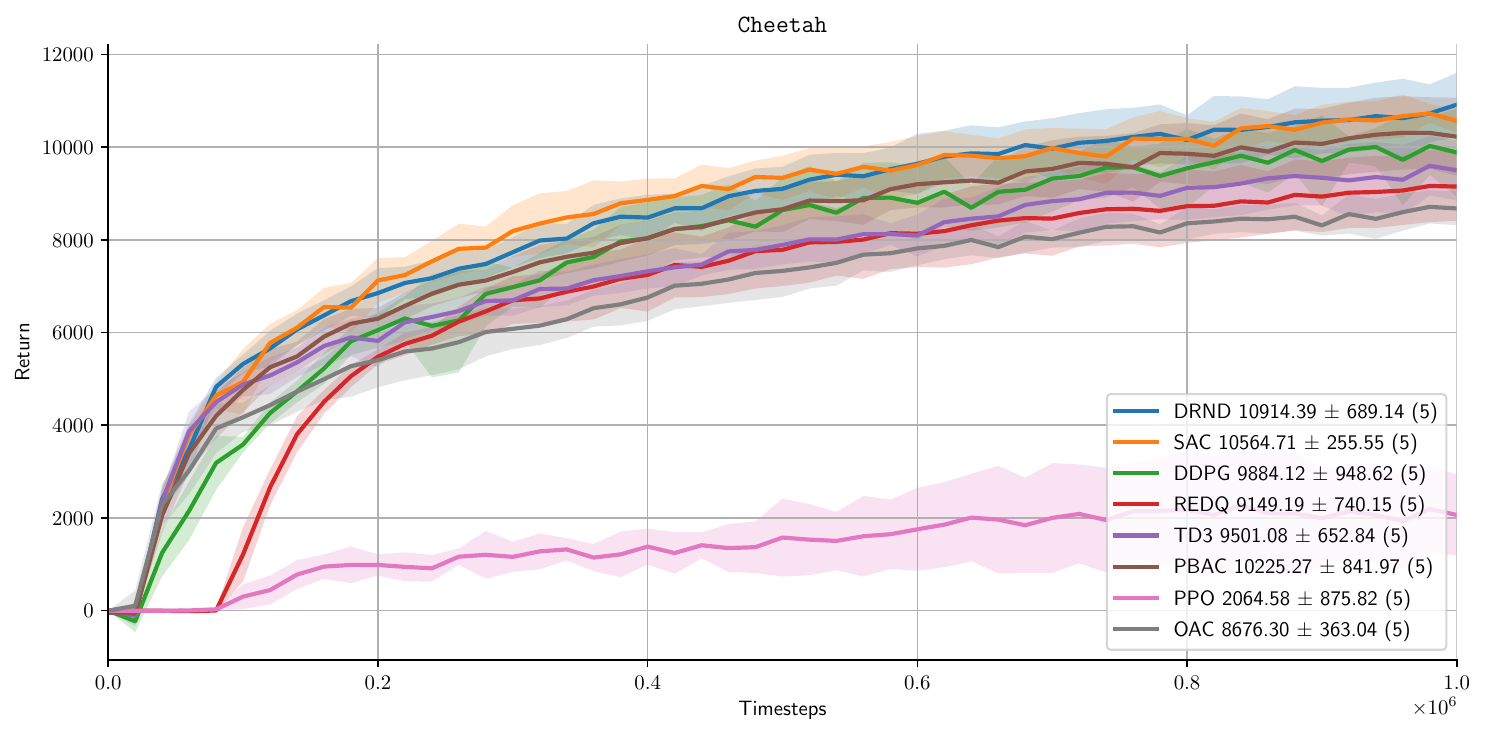}
        \label{fig:cheetah}
    \end{subfigure}
    \begin{subfigure}[b]{0.45\textwidth}
        \centering
        \includegraphics[width=\textwidth]{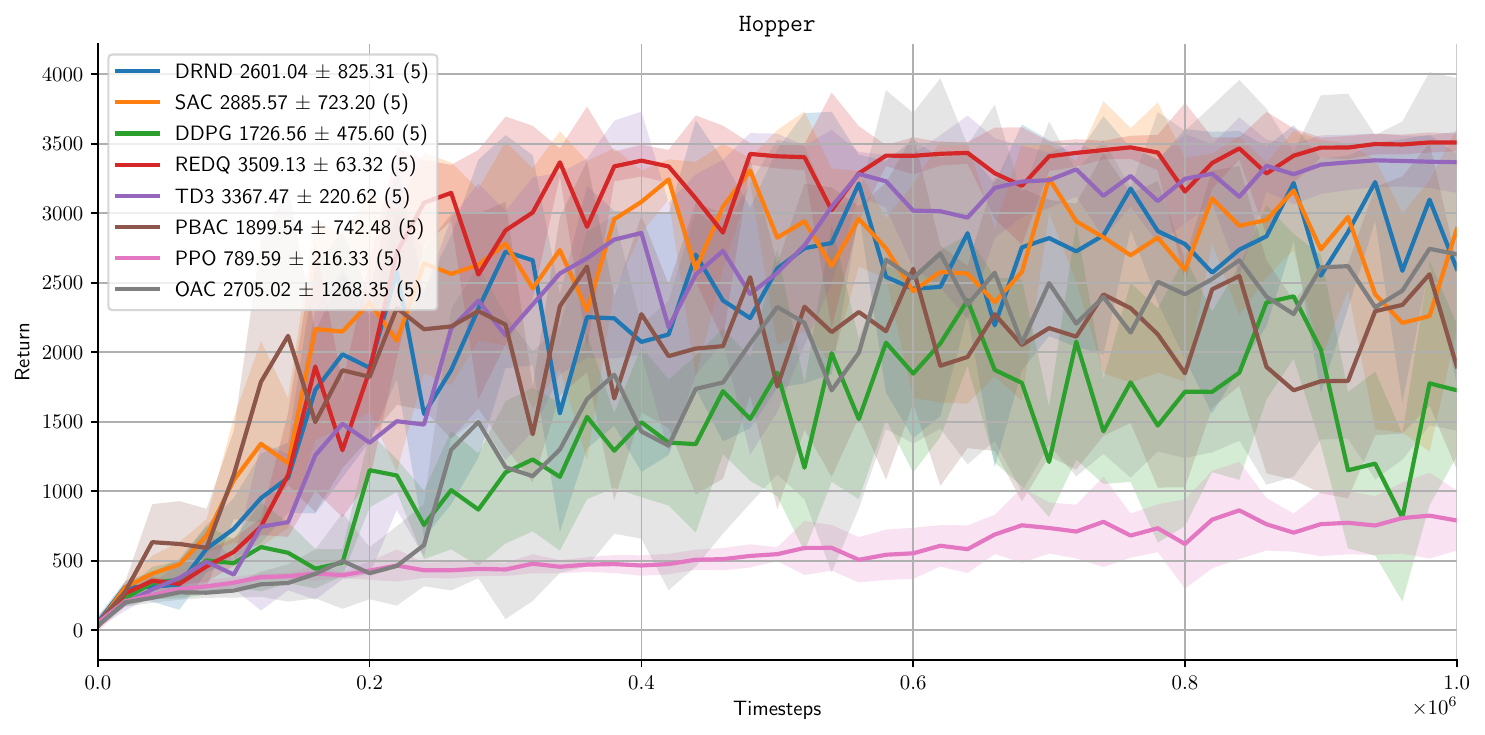}
        \label{fig:hopper}
    \end{subfigure}
    \hfill
    \begin{subfigure}[b]{0.45\textwidth}
        \centering
        \includegraphics[width=\textwidth]{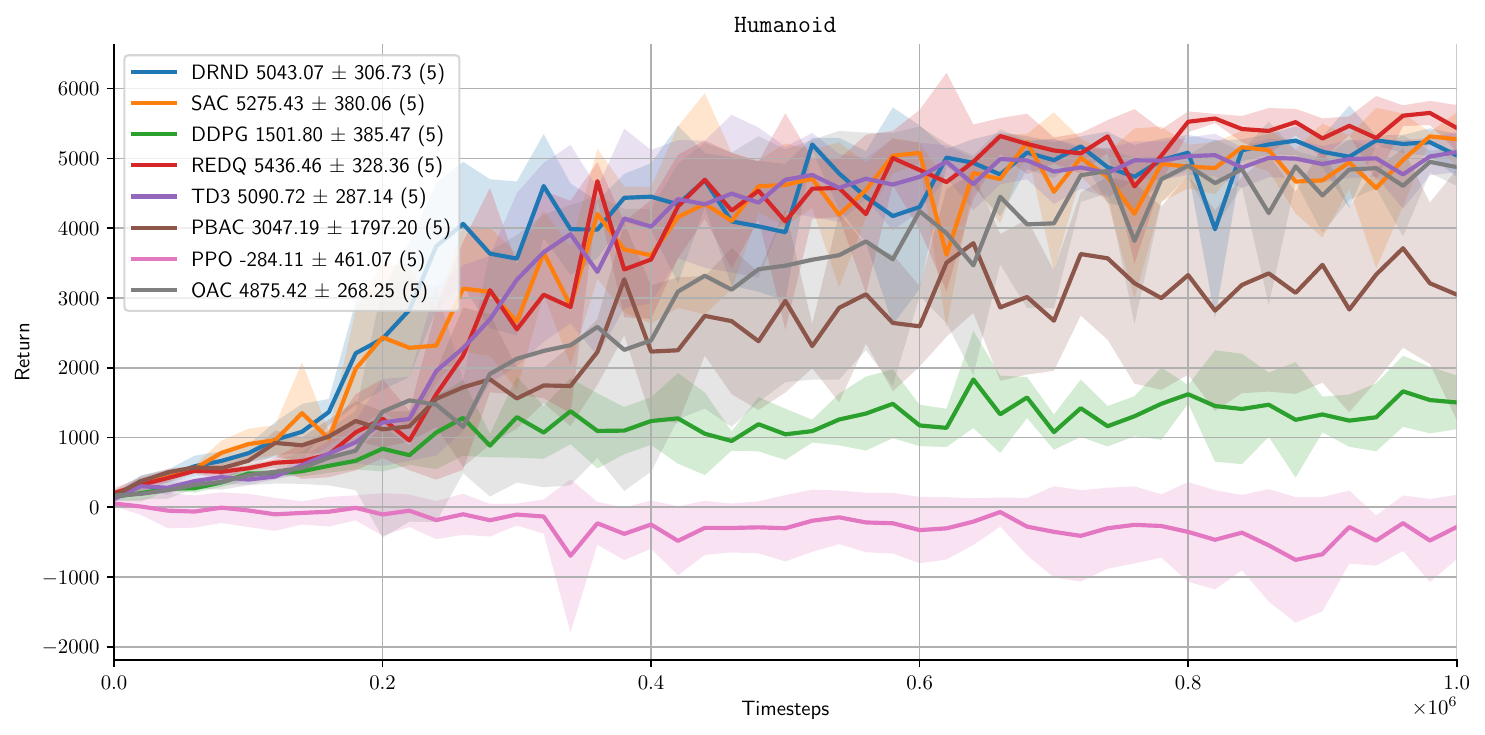}
        \label{fig:humanoid}
    \end{subfigure}
    \begin{subfigure}[b]{0.5\textwidth}
        \centering
        \includegraphics[width=\textwidth]{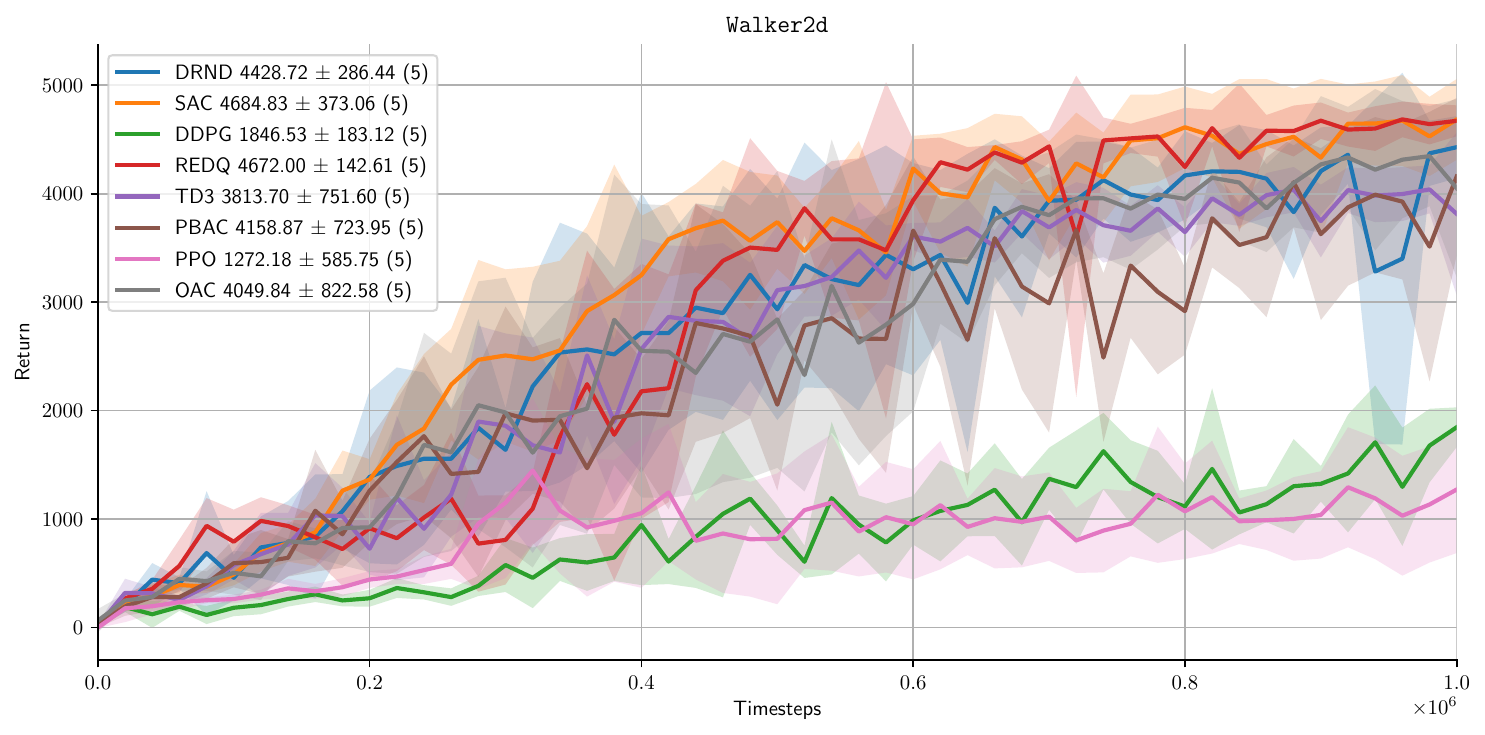}
        \label{fig:walker}
    \end{subfigure}
    \caption{Evaluation results on MuJoCo environments.}
    \label{fig:eval_results}
\end{figure}

Our experiments use environments compatible with the Gymnasium interface \citep{towers2024gymnasium}. In~\cref{fig:eval_results}, we report results for five standard continuous control environments from the MuJoCo suite \citep{todorov2012mujoco} which are frequently used in the literature for benchmarking. Integration of other environments into our codebase is straightforward via custom wrappers that conform to the Gymnasium API.

\section{Contributions}

GB: Organization, implementation, documentation, writing.
AA: Implementation, documentation, evaluation, writing.
MH: Implementation, documentation, evaluation, writing.
BT: Implementation, evaluation.
NW: Writing.
YSW: Documentation.
MK: Organization, implementation, ideation, writing.

\vskip 0.2in
\bibliography{references}

\end{document}